\documentclass[prologue,dvipsnames,svgnames,sigconf,nonacm]{acmart}

\usepackage{xcolor}
\usepackage{amsmath,amsfonts} 
\usepackage[normalem]{ulem}
\usepackage{xspace}
\usepackage{graphicx}
\usepackage{alltt}
\usepackage{color}
\usepackage{subfig}
\usepackage{comment}
\usepackage{hyperref}
\hypersetup{
    colorlinks=true,
    citecolor=black,
    linkcolor=black,
    urlcolor=blue
}

\newcommand{\pyarg}{\textsc{PyArg}\xspace}
\newcommand{\afxray}{\textsc{AF-Xray}\xspace}
\newcommand{\xray}{\textsc{Xray}\xspace}
\newcommand{\mypara}[1]{\textbf{#1.}}
\newcommand{\IN}{\textsf{\mbox{\textsc{in}}}\xspace}
\newcommand{\OUT}{\textsf{\mbox{\textsc{out}}}\xspace}
\newcommand{\UNDEC}{\textsf{\mbox{\textsc{undec}}}\xspace}

\newcommand{\figref}[1]{Fig.\,\ref{#1}}
\newcommand{\Figref}[1]{Figure\,\ref{#1}}
\newcommand{\pos}[1]{\textsf{#1}}

\sloppypar
\setcopyright{acmcopyright}
\copyrightyear{2025}
\acmYear{2025}
\setcopyright{rightsretained}
\acmConference[ICAIL 2025]{20th Intl.\ Conf.\  on AI and Law}{June 16--20, 2025}{Chicago, IL, USA}
\acmDOI{}
\acmISBN{XXX}

\begin{document}

\title{\afxray: Visual Explanation and Resolution of Ambiguity in Legal Argumentation Frameworks}

\author{Yilin Xia}
\affiliation{
  \institution{University\ of Illinois, Urbana-Champaign}
  \city{\relax} \state{\relax}
  \country{\relax}
}
\email{yilinx2@illinois.edu}

\author{Heng Zheng}
\affiliation{
  \institution{University\ of Illinois, Urbana-Champaign}
  \city{\relax}  \state{\relax}
  \country{\relax}
}
\email{zhenghz@illinois.edu}

\author{Shawn Bowers}
\affiliation{
  \institution{Gonzaga University, Spokane, WA}
  \city{\relax}  \state{\relax}
  \country{\relax}
}
\email{bowers@gonzaga.edu}

\author{Bertram Lud\"{a}scher}
\affiliation{
  \institution{University\ of Illinois, Urbana-Champaign}
  \city{\relax}   \state{\relax}  \country{\relax}
}
\email{ludaesch@illinois.edu}

\renewcommand{\shortauthors}{Y.\ Xia, H.\ Zheng, S.\ Bowers, B.\ Lud\"{a}scher}

\maketitle

\section{Introduction}
Abstract argumentation frameworks \cite{dung1995acceptability} offer well-established, formal approaches for representing and reasoning about case law~\cite{BenchCapon20}. Given an argument $x$ in an argumentation framework (AF), it is easy to determine the status of $x$ under {skeptical} reasoning, i.e., whether $x$ is \emph{accepted} (\IN), \emph{defeated} (\OUT), or \emph{undecided} (\UNDEC).  In case of the latter, the AF is \emph{ambiguous}: it has a 3-valued grounded semantics $S_0$, and some conflicts may require additional assumptions or choices to be made to resolve these ambiguities. \emph{Value-based} 
and \emph{Extended}
AFs have been  used in legal reasoning  to resolve and justify the acceptance in such scenarios~\cite{BenchCaponModgil09}. These approaches help users explain choices among alternative resolutions by discounting (or ignoring) certain attack edges, e.g., based on social value preferences.
For AF non-experts, however, it can be difficult to pinpoint the specific reasons (i.e., \emph{critical attacks}) causing an ambiguity, and to visualize an AF's semantics in a way that all parties understand. 

We present \afxray\footnote{\href{https://github.com/idaks/xray}{\afxray}: \textbf{A}rgumentation \textbf{F}ramework e\textbf{\underline{X}}planation, \textbf{\underline{R}}easoning, and \textbf{\underline{A}}nal\textbf{\underline{Y}}sis \cite{AFXRayRepo2025}}, a novel platform for exploring, analyzing, and visualizing AF solutions, which builds upon the state-of-the-art open source \pyarg system~\cite{odekerken2023pyarg}. \xray ``looks deeper'' into the structure of AFs and provides new analysis and visualization components for explaining the acceptance of arguments under \emph{skeptical} reasoning, and for identifying {critical attacks}, whose suspension \emph{resolves} undecided arguments under \emph{credulous} reasoning. It adds: 

$(i)$ A novel \emph{layered} AF visualization, based on the game-theoretic \emph{length}\footnote{$\dots$ which is closely related to   \emph{min-max numberings}  of \emph{strongly admissible sets}~\cite{caminada2019strong}} (or \emph{remoteness}~\cite{smith_graphs_1966}) of nodes~\cite{BowersXiaLudaescherSAFA24}; 
$(ii)$ a novel \emph{classification of attack edges} derived from their game-theoretic type~\cite{BowersXiaLudaescherSAFA24};
$(iii)$ the ability to switch between alternate 2-valued solutions $S_1,\dots,S_n$ of an AF (visualized as \emph{overlays} on the ambiguous, 3-valued $S_0$); and 
$(iv)$ the identification and display of \emph{critical attacks} in $\boldsymbol{\Delta}_{i}$ for each 
solution $S_i$, where $\boldsymbol{\Delta}_i =\{\Delta_{i,1}, \dots, \Delta_{i,n_i}\}$ are $n_i$ critical attack sets $\Delta_{i,j}$ for $S_i$: Temporarily suspending the attacks in $\Delta_{i,j}$ yields a 2-valued grounded solution $S'_{i,j}$. Together, the suspension of $\Delta_{i,j}$ and the resolution $S'_{i,j}$ explain the choices for $S_i$.

\section{\afxray in Action}

In \xray, similar to \pyarg, users input  AFs as graphs $G = (V, E)$ of arguments $V$ and attack edges $E \subseteq V \times V$, either via a web interface or file upload. The input graph is then  visualized. Users pick a semantics (e.g., \emph{grounded}, \emph{stable}, \emph{preferred}) and select one of the possible \emph{solutions} (labelings). Arguments are colored according to their status: \IN{} (blue), \OUT{} (orange), and \UNDEC{} (yellow). 
The following highlight some of \xray's features.



\begin{figure*}
  \centering
  \subfloat[Layered \emph{skeptical} (grounded) solution $S_0$]{
    \includegraphics[width=.29\textwidth]{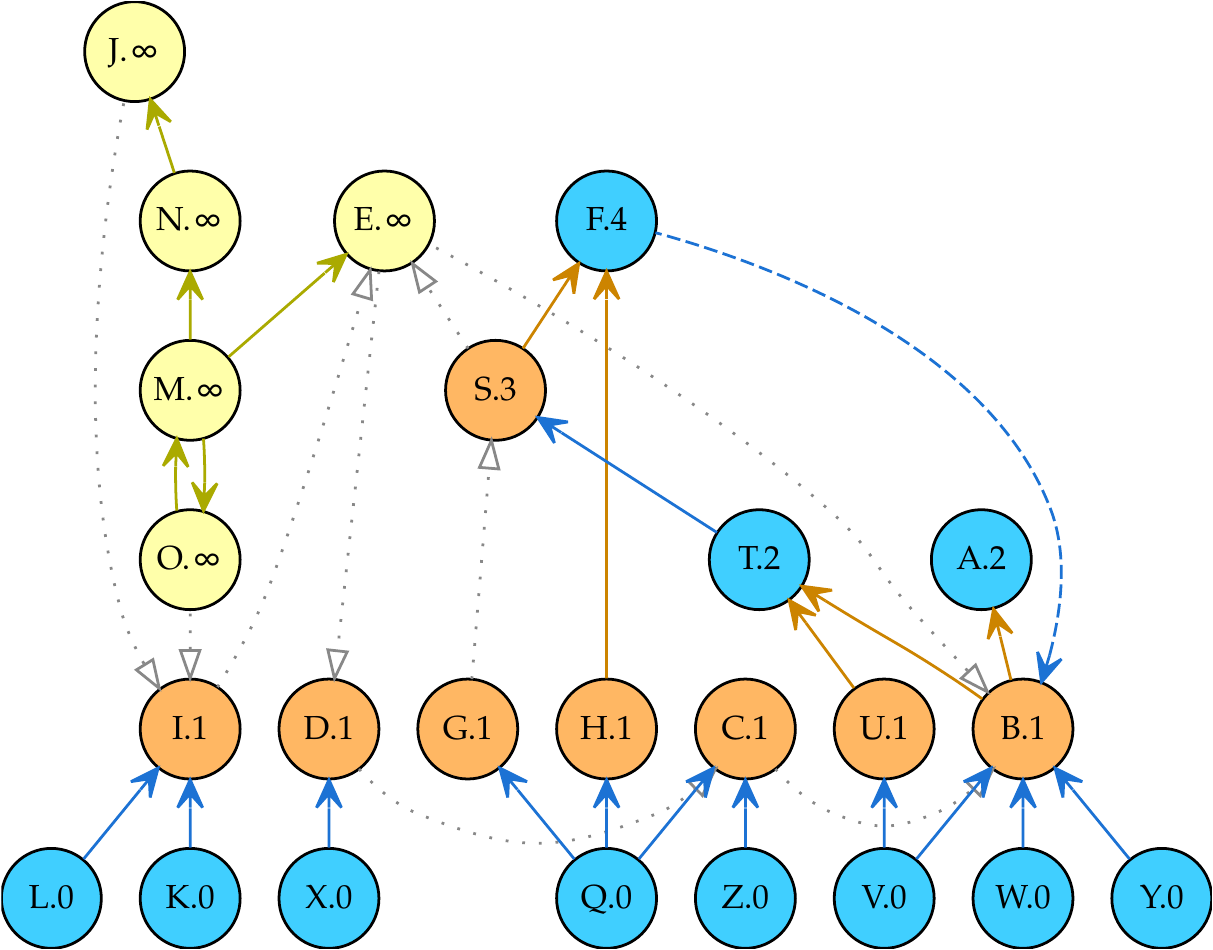}
    \label{fig:layered}}
\hspace{.045\textwidth}
  \subfloat[Overlay \emph{credulous} (stable) resolution $S'_{1,1}$]{
    \includegraphics[width=.29\textwidth]{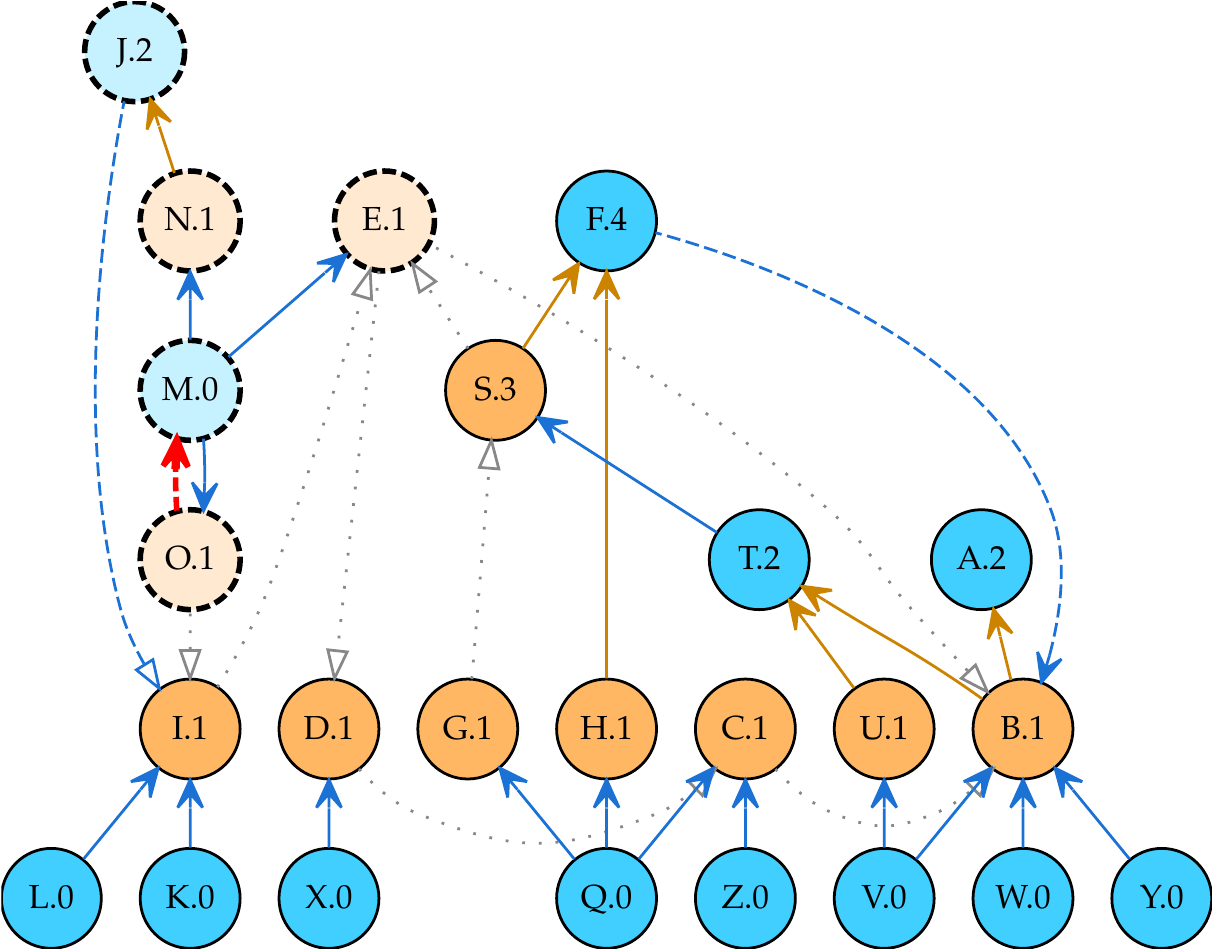}
    \label{fig:stable-1}}
  \hspace{.045\textwidth}
  \subfloat[Overlay \emph{credulous} (stable) resolution $S'_{2,1}$]{
    \includegraphics[width=.29\textwidth]{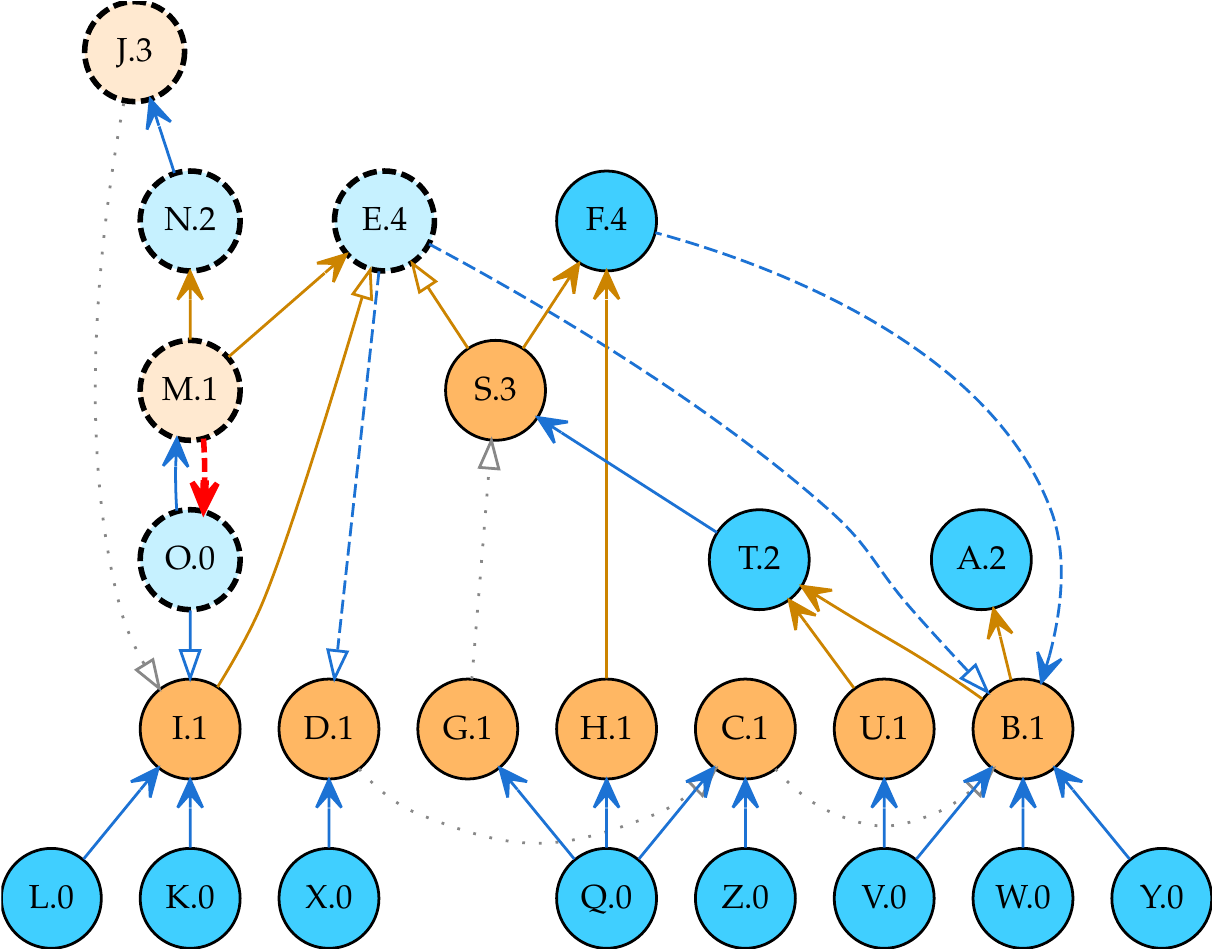}
    \label{fig:stable-2}}
 \caption{\small \afxray visualizations of the \emph{Wild Animals} cases \cite{bench-capon_representation_2002}: (a) The ambiguous (3-valued) grounded solution $S_0$ uses the  \emph{length} of nodes: e.g., \textsf{F.4} requires no more than four discussion rounds to prove that \pos F is \IN.
 Distinct \emph{edge types} are used to account for their semantic roles~\cite{BowersXiaLudaescherSAFA24}. The  \emph{overlays} in (b) and (c) represent {alternative} {resolutions} $S'_{1,1}$ and $S'_{2,1}$: The \UNDEC nodes \pos E, \pos J, \pos M, \pos N, \pos O  in (a) have been \emph{decided} (\pos M is \IN and \pos O is \OUT in $S_1$; and vice versa in $S_2$). These choices are {explained} by \emph{critical attacks} (red edges) $\boldsymbol{\Delta}_1=\{\Delta_{1,1}{:}\,\{\pos O\to \pos M \}\}$ and
 $\boldsymbol{\Delta}_2=\{\Delta_{2,1}{:}\,\{\pos M\to \pos{O} \}\}$, i.e., minimal sets of (temporarily) \emph{suspended} edges: when suspensions are applied,  2-valued  grounded solutions $S'_{1,1}$, $S'_{2,1}$ are obtained for $S_1$ and $S_2$.
}
 \label{fig:3panel}
\end{figure*}

\mypara{Layered Visualizations} \figref{fig:layered} shows the \emph{Wild Animals} legal example using 
\xray's layered visualization. The layering is based on the \emph{length} of argument nodes, which can be computed alongside the  grounded labeling $S_0$~\cite{van1993alternating,BowersXiaLudaescherTAPP24,BowersXiaLudaescherSAFA24}. 
In $S_0$, an argument $x$ that is  \OUT has an \IN-labeled attacker; $x$ is \IN if every attacker of $x$ is \OUT; and \UNDEC if $x$ is neither \IN nor \OUT in $S_0$. In the layered visualization, the bottom layer consists of  arguments that are trivially labeled \IN because they have no attackers (length\,=\,0); the next layer consists of \OUT arguments (length\,=\,1) that are defeated by length-0 attackers, etc. \UNDEC arguments result from unfounded attack-chains (length\,=\,``$\infty$''), and are displayed outside the layering. Arguments that {justify} an \IN or \OUT-labeled argument $x$ are located at layers below $x$: e.g.,  while \pos{F.4} in \figref{fig:layered} attacks \pos{B.1}, the defeat of \pos{B.1} is known (due to \pos{V.0}, \pos{W.0}, and \pos{Y.0}) \emph{before} \pos{F.4}'s label is determined. The layering makes the \emph{well-founded} (and thus ``self-explanatory") derivation structure of the grounded semantics explicit. 

\mypara{Visualizing Attack Types} \xray visualizes attacks according to their role in determining argument labels \cite{BowersXiaLudaescherSAFA24}. Successful (blue) attacks are classified as either \emph{primary} (solid blue) or \emph{secondary} (dashed blue). Secondary attacks point to arguments with smaller lengths, e.g., \pos F's attack on \pos B, whose defeat was established in a lower layer. Dotted gray edges are ``\emph{blunders}'', i.e., an edge type which is  irrelevant for the acceptance status (\emph{provenance}) of arguments \cite{BowersXiaLudaescherSAFA24}. A minimal explanation of an argument excludes secondary attacks and blunders, so they are de-emphasized in the visualization.    

\mypara{Resolving Ambiguity} To analyze and disambiguate the \UNDEC portion of a 3-valued grounded solution, a less skeptical 2-valued semantics (e.g., stable or preferred) can be employed by \xray to  enumerate these alternative solutions.
Each solution represents a choice for resolving the (direct or indirect) circular conflicts that created the ambiguities (\UNDEC nodes in  \figref{fig:layered}) in the first place.
The two solutions  in \figref{fig:stable-1}\,\&\,\ref{fig:stable-2} are depicted as hybrid \emph{overlays} of the 3-valued grounded solution $S_0$  (with \UNDEC nodes) and the respective 2-valued stable solution $S_i$ (without \UNDEC nodes):  The colors (\IN/\OUT-labels) of the stable solutions $S_i$ are visualized ``on top of'' the grounded solution $S_0$, i.e., they share the same layered visualization, but now with \UNDEC arguments colored according to their (newly resolved) acceptance status in $S'_i$. In such overlays, lighter colors and dashed outlines mark the original \UNDEC subgraph. 

\mypara{Explaining Credulous Solutions in \xray} The grounded solution $S_0$ of an AF (\figref{fig:layered}) is self-explanatory: \IN, \OUT, and \UNDEC arguments are justified by their well-founded derivation and the \emph{length}\footnote{The node length can be computed as a by-product of computing the well-founded model via the \emph{alternating fixpoint procedure}~\cite{van1993alternating}.} used to rank nodes in the layered visualization \cite{BowersXiaLudaescherTAPP24,xia_layered_vis_demo_2024}. 
The explanatory structure of credulous (e.g., stable) solutions is more complex, however. It consists of a well-founded part (blue/orange nodes in \Figref{fig:3panel}) and an ambiguous part (yellow nodes in \figref{fig:layered}). 
A large number of alternative 2-valued solutions $S_i$ usually ``hide'' in the ambiguous parts of $S_0$. In \xray, these choices  can be explained via sets of \emph{critical attacks} $\Delta_{i,j}$.  
If we choose to suspend these minimal sets of edges  (e.g., via temporary deletions), every previously \UNDEC argument $x$ will be either \IN or \OUT, and for the chosen suspension $\Delta_{i,j}$, 
there is now a well-founded derivation of $x$. In this way, \xray allows the user to pinpoint critical  attacks and arguments to support a desired outcome within the confines of the initial grounded solution. This approach facilitates new use cases for legal reasoning that complement earlier approaches such as Value-based and Extended AFs \cite{BenchCaponModgil09}. Whereas the latter assume that users already know which edges to attack, \xray systematically generates all such sets of critical edges, thus providing a deeper semantic analysis than any state-of-the-art system we are aware of.

\mypara{Demonstration Overview} 
The \href{https://github.com/idaks/xray}{demonstration} will illustrate: (1)~loading an AF with legal annotations of abstract arguments; (2)~the layered visualization of the grounded solution $S_0$, observing the well-founded derivations of arguments (\figref{fig:layered}); (3)~exploration of different stable solutions $S_i$ and their overlays $S'_{i,j}$, observing critical attack sets that explain the choices (suspensions) made (\figref{fig:stable-1}, \ref{fig:stable-2}) as part of the resolution; and (4) exporting the desired (re)solutions for future use.
Legal argument annotations (hovering over a node displays its annotation; clicking on it navigates to a page with details) are used to discuss a real-world example: We study the mutual attack between two arguments: \pos M  (\emph{mere pursuit is not enough}) and \pos O (\emph{bodily seizure is not necessary}), which directly reflects opposing arguments in \emph{Pierson v.~Post}~\cite{bench-capon_representation_2002}. 
Users can toggle between stable solutions $S_1$ and $S_2$ and view the critical attack sets $\boldsymbol{\Delta}_1=\{\Delta_{1,1}\}$ and
 $\boldsymbol{\Delta}_2=\{\Delta_{2,1}\}$
explaining each legal \emph{possible world}.
This supports the teleological structure of legal reasoning: Different assumptions lead to different legally justified conclusions, e.g., depending on which social values are prioritized~\cite{bermanRepresentingTeleologicalStructure1993a}.

\bibliographystyle{ACM-Reference-Format}
\bibliography{icail}

\end{document}